\relax
\documentclass[10pt,twocolumn]{article}
\usepackage{simpleConference}  
\usepackage{times}  
\usepackage{helvet}  
\usepackage{courier}  
\usepackage{url}  
\usepackage{graphicx}  
\usepackage{color}
\usepackage{bbm}
\usepackage{verbatim}
\usepackage[numbers]{natbib}

\usepackage{multirow}

\usepackage{amsmath, amssymb, amsfonts, amsthm}
\usepackage{algorithm, algorithmic}
\newtheorem{theorem}{Theorem}
\newtheorem{lemma}[theorem]{Lemma}
\theoremstyle{definition}
\newtheorem{definition}{Definition}

\theoremstyle{assumption}

\DeclareMathOperator{\E}{\mathbb{E}}

\DeclareMathOperator{\R}{\mathbb{R}}

\DeclareMathOperator{\YY}{\mathcal{Y}}

\newcommand{\Ind}[1]{\mathbbm{1}_{#1}}

\DeclareMathOperator*{\argmax}{argmax}

\begin{document}
\title{
MEMOIR: Multi-class Extreme Classification with Inexact Margin}
\author{Anton Belyy$^1$, Aleksei Sholokhov$^2$ \\
ITMO University$^1$, University of Washington$^2$ \\
anton.belyy@gmail.com$^1$, aksh@uw.edu$^2$ 
}


\maketitle
\begin{abstract}
Multi-class classification with a very large number of classes, or extreme classification, is a challenging problem from both statistical and computational perspectives. Most of the classical approaches to multi-class classification, including one-vs-rest or multi-class support vector machines, require the exact estimation of the classifier's margin, at both the training and the prediction steps making them intractable in extreme classification scenarios. In this paper, we study the impact of computing an approximate margin using nearest neighbor (ANN) search structures combined with locality-sensitive hashing (LSH). This approximation allows to dramatically reduce both the training and the prediction time without a significant loss in performance. We theoretically prove that this approximation does not lead to a significant loss of the risk of the model and provide empirical evidence over five publicly available large scale datasets, showing that the proposed approach is highly competitive with respect to state-of-the-art approaches on time, memory and performance measures.
\end{abstract}

\section{Introduction}

Recently, the problem of large-scale multi-class classification became very popular in the machine learning community owing to numerous applications in computer vision \cite{perronnin2007fisher,joulin2012multi}, text categorization \cite{joachims1998text}, recommendation and ranking systems \cite{song2008real,shashua2003ranking}. Publicly available text and image repositories such as Wikipedia, Yahoo! Directory\footnote{\url{www.dir.yahoo.com}}, ImageNet\footnote{\url{www.image-net.org}} or Directory Mozilla DMOZ \footnote{\url{www.dmoz.org}} contain millions of objects from thousands of classes. For instance, an LSHTC dataset from Mozilla DMOZ collection contains 163K objects belonging to 12K classes and described by 409K features. Classical multi-class classification approaches, such as one-vs-one and one-vs-rest, remain popular mainly because of their high accuracy and robustness to noisy input. On the other hand, their direct application to the extreme classification problems is doubtful due to highly non-linear time and memory efforts \cite{read2011classifier,Shalev-Shwartz2011}.

Some promising attempts have been made to reduce the computation time of these models, by either doing locality-sensitive hashing \cite{neyshabur2014symmetric} or reinforcement learning policies in an online convex optimization context \cite{auer2002finite}. Despite empirical evidence of how well these approaches perform, to the best of our knowledge none of the studies propose  a well funded theoretical strategy for large-scale multi-class classification that guarantees a gain in computation time without a significant loss on the statistical risk of the model.

In this paper, we propose a novel method for approximating the output estimation of multi-class classification models using approximate nearest neighbor (ANN) search structures with locality-sensitive hashing (LSH). We show that the proposed inexact margin computation significantly reduces time and memory requirements, allowing popular techniques, such as multi-class support vector machines and the one-vs-rest approach, to pass the scale without a significant loss of their true risk.
 
The contribution of this paper is threefold. Namely, we 
\begin{enumerate}
\item propose an inexact margin multi-class classification framework and provide a theoretical analysis of its behavior; 
\item design efficient numerical methods for $\ell_1$ and $\ell_2$ inexact margin multi-class support vector machines; 
\item provide empirical evidence of its ability to learn efficient models compared to state-of-the-art approaches over multiple extreme classification datasets  and make available the corresponding open-source software for research purpose. 
\end{enumerate}

In the next section we introduce a framework of multi-class margin classification and describe the inexact margin idea in more detail. Then we provide a theoretical analysis of the statistical performance of multi-class classification methods with inexact margin supporting it by the excess risk bounds of the corresponding classification algorithms. Further, we present experimental results obtained on publicly available extreme classification benchmarks, and finally, we briefly discuss the proposed algorithms and compare them to existing solutions.
\section{Multi-class Margin Classification}

Let $S = \{(x_i, y_i)\}_{i=1}^n$ be an identically and independently distributed (i.i.d.) sample with respect to a fixed yet unknown distribution ${\cal D}$, over ${\cal X}\times {\cal Y}$, where ${\cal X}\subseteq {\mathbb R}^d$ is a feature space and ${\cal Y}$, $|{\cal Y}| = C < \infty$ is a set of classes. Given a hypothesis set ${\cal F}$ of functions mapping ${\cal X}\times {\cal Y}$ to $\mathbb{R}$, the exact margin of a labeled example $(x, y)$ with respect to a function $f \in {\cal F}$ is defined as 
\begin{gather}\label{margin}
m_f(x, y) = f(x, y) - \max_{{y' \in {\cal Y}, y'\neq y}} f(x, y'). 
\end{gather}
An observation $(x, y)$ is misclassified by a function $m_f\in{\cal F}$ if and only if $m_f(x, y)~\le~0$. We refer to a class of margin loss functions as 
\[{\cal M} = \{m:\; m(x,y) = f(x, y) - \max_{y'\neq y} f(x, y), f(x, \cdot)\in {\cal F}\}.\] 
The main problem here is that the computation of the margin for an observation requires the estimation of $f(x, y)$ for each $y \in {\cal Y}$ which is intractable when $|{\cal Y}| = C>\!\!>1$ is too large. For instance, in the case of linear classifiers, margin computation is equal to finding the maximal element of a matrix-vector product on each iteration which is challenging in large-scale scenarios. In order to overcome this problem in such case, we estimate an approximate, or an inexact margin for each observation $(x, y)$ by first choosing randomly a class $y'\in {\cal Y}, y'\neq y$ and then estimating 
\begin{equation}
\label{appmargin}
{\bar m}_f(x, y)= f(x, y) - f(x,y').
\end{equation}

In this paper, we focus on the influence of inexact margin computation over the true risk of a classifier $f\in~{\cal F}$
\[
 	R(f) = \mathbb{E}_{(x, y) \sim {\cal D}}[\Ind{m_f(x, y) \le 0}], 
\]
where $\Ind{\pi}$ is equal to $1$ if the predicate $\pi$ is true and $0$ otherwise. More precisely, we are interested in the case where the classifier is found following the Empirical Risk Minimization principle by supposing that the approximate margin ${\bar m}_f$ is \emph{  $(\varepsilon, \delta)$ inexact}, that is for a given $(x, y) \sim {\cal D}$ and with probability at least $1-\delta$, $\delta > 0$,  we have 
\[
	m_f(x, y) - {\bar m}_f(x,y) \le \varepsilon.
\]
The empirical loss considered in this work is based on the Hinge $\rho$-loss function $\ell(m_f(x, y)) = (1 - \rho^{-1} m_f(x,y))_+$, (resp. $\ell(\bar m_f(x, y))$) for $\rho>0$ and defined as
\[
\widehat{R}_{\rho, n}(m_f) = \frac{1}{n}\sum_{i=1}^n \ell(m_f(x, y)).
\]

Our main result is stated in Theorem \ref{thm:rad}, and it provides an upper-bound on the true risk of a multi-class classifier based on its empirical loss estimated with an \emph{  $(\varepsilon, \delta)$ inexact} margin. The notion of  function  class  capacity  used  in  the  bound is the Rademacher complexity $\mathfrak{R}({\cal H})$ of the set of functions  $\mathcal{H}=\{x\mapsto f(x,C), f\in\mathcal{F}\}$ \cite{mohri2012foundations}: 
\[
    \widehat{\mathfrak{R}}_n({\cal H}) = 
    \mathbb{E}_{\sigma} \sup_{f \in {\cal F}} \frac{1}{n} \sum_{i=1}^n \sigma_i f(x_i,C), \;       \mathfrak{R}_n({\cal H}) = \mathbb{E}_{S}\widehat{\mathfrak{R}}_n({\cal H}) 
\]

where $\sigma_i$'s are independent uniform random variables taking values in $\{-1,+1\}$; i.e. $\forall i, \mathbb{P}(\sigma_i=~-~1)=~\mathbb{P}(\sigma_i=~1)=~\frac{1}{2}$.
\begin{theorem}\label{thm:rad}
Let $S = \{(x_i, y_i)\}_{i=1}^n$ be an i.i.d. sample from an unknown distribution ${\cal D}$ over ${\cal X}\times {\cal Y}$, $|{\cal Y}| = C$, and ${\cal F}$ be a class functions from ${\cal X}\times {\cal Y} \to \mathbb{R}$. Then for any $\rho > 0$ with probability at least $1 - \delta$, the expected risk of any $f \in {\cal F}$ trained with a $(\varepsilon \rho, \delta/n)$ inexact margin is upper-bounded by 
\begin{gather}\label{eq:rad-bound}
\hspace{-1mm}R(f) \le {\widehat R}_{\rho(1+\varepsilon), n}({\bar m}_f) + \frac{4 C}{\rho} \widehat{\mathfrak{R}}_n({\cal H}) + 3 \sqrt{\frac{\log 4/\delta}{2n}}. 
\end{gather}

Moreover, for kernel-based hypotheses, with $K:\mathcal{X}\times\mathcal{X}\rightarrow \mathbb{R}_+$ a PDS kernel and $\Phi:\mathcal{X}\rightarrow \mathbb{H}$ its associated feature mapping function, defined as :
\[
\mathcal F_\Omega=\{f:(x,y)\in\mathcal{X}\times\mathcal{Y}\mapsto \phi(x)^\top \omega_y\mid  \|\mathbf{W}\|_2 \le \Omega\}
\]
where $\mathbf{W}=(\omega_1\ldots,\omega_{C})$ is the matrix formed by the $C$ weight vectors defining the kernel-based hypotheses,  and $||\mathbf{W}||_{\mathbb{H}}=\left(\sum_{y\in \cal{Y}}||\omega_y||^2\right)^{1/2}$ is the $L_{\mathbb{H}}^2$ group norm of  $\mathbf{W}$, then if $\forall x, \|\phi(x)\|_2 \le R$ one has
\begin{gather}\label{eq:rad-bound-kernel}
\hspace{-1mm} R(f) \le \widehat{R}_{\rho(1+\varepsilon), n}({\bar m}_f) + \frac{4 C}{\rho} \sqrt{\frac{R^2 \Omega^2}{n}} + 3 \sqrt{\frac{\log 4/\delta}{2n}}. 
\end{gather}
\end{theorem}

\begin{proof}
The standard Rademacher complexity margin bound according to Theorem 4.4 of \cite{mohri2012foundations} gives with probability $1-\delta/2$ for a class of functions ${\cal M}$ the following bounds: 
\begin{gather}\label{eq:mr1}
R(f) \le {\widehat R}_\rho({m}_f) + \frac{2}{\rho} {\mathfrak{R}}_n({\cal M}) +  \sqrt{\frac{\log 2/\delta}{2n}},
\end{gather}
and
\begin{gather}\label{eq:mr2}
R(f) \le {\widehat R}_\rho({m}_f) + \frac{2}{\rho} \widehat{\mathfrak{R}}_n({\cal M}) + 3 \sqrt{\frac{\log 4/\delta}{2n}},
\end{gather}
where $\widehat{R}_{\rho, n}({\bar m}_f) = \frac{1}{n}\sum_{i=1}^n (1 - \rho^{-1} {\bar m}_f(x_i, y_i))$. 


Similarly to \cite{lei2015multi}, for all $i$, let  $z_i\in\mathcal{Y} $ satisfy ${m}_f(x_i, y_i) = f(x_i, y_i) - f(x, z_i)$ we have due to monotonicity of the Rademacher complexity in the number of examples
\begin{align}\label{eq:rad-k}
&{\mathfrak{R}}_n({\cal M}) = \mathbb{E}_{\sigma} \sup_{f \in {\cal F}} \frac{1}{n} \sum_{i=1}^n \sigma_i {m}_f(x_i, y_i)  \le \nonumber\\
&\;\;\mathbb{E}_{\sigma} \sup_{f \in {\cal F}} \frac{1}{n} \sum_{i: y_i, z_i\neq C} \sigma_i f(x_i, y_i) + \nonumber\\
&\;\;\mathbb{E}_{\sigma}  \sup_{f \in {\cal F}}  \frac{1}{n} \hspace{-1mm}\left\{\sum_{i: y_i =  C} \sigma_i f(x_i, C) \hspace{-1mm} + \hspace{-1mm}\sum_{i: z_i =  C} \hspace{-1mm} - \sigma_i f(x_i, C)\right\}\le \nonumber\\
& \;\;\mathbb{E}_{\sigma}  \sup_{f \in {\cal F}}  \frac{1}{n} \hspace{-1mm}\sum_{i: y_i, z_i\neq C} \sigma_i f(x_i, y_i)+ 2 \cdot \widehat{\mathfrak{R}}_n({\cal H}) \le 2C  \widehat{\mathfrak{R}}_n({\cal H})
\end{align}

Finally, with probability at least $1-\delta/2$ in the conditions of the theorem, for all training objects $\{(x_i, y_i)\}_{i=1}^n$ we have $m_f(x_i,y_i) - {\bar m}_f(x_i, y_i) \le \varepsilon \rho$, thus 
$\widehat{R}_{\rho, n}(m_f) \le {\bar R}_{\rho(1+\varepsilon), n}({\bar m}_f).$

Combining it with Ineq.~\eqref{eq:rad-k} one gets 
\begin{gather}\label{eq:1-pre-fin}
R(f) \le {\widehat R}_\rho({\bar m}_f) + \frac{4C}{\rho} \widehat{\mathfrak{R}}_n({\cal H}) + 3 \sqrt{\frac{\log 4/\delta}{2n}}.
\end{gather}

Application of Ineq.~\eqref{eq:1-pre-fin} to Ineq.\eqref{eq:mr2} proves Ineq.~\eqref{eq:rad-bound} in the statement of the theorem. 
Theorem 4.3. of \cite{mohri2012foundations} gives for the linear classifiers 
$
\mathfrak{R}({\cal H}) \le \sqrt{R^2\Omega^2/n}
$
which proves the Ineq.~\eqref{eq:rad-bound-kernel}.  

To proof the remaining inequalities we note that with probability at least $1-\delta/2$ in the conditions of the theorem, for all training objects $\{(x_i, y_i)\}_{i=1}^n$ we have $m_f(x_i,y_i) - {\bar m}_f(x_i, y_i) \le \varepsilon \rho$, thus 
$
\widehat{R}_{\rho, n}(m_f) \le {\bar R}_{\rho(1+\varepsilon), n}({\bar m}_f).
$ 
\end{proof}



\paragraph{Margin approximation.} We consider two principal approaches to inexact margin computation: locality-sensitive hashing and convex optimization. Our main focus here is linear classification and the maximal inner product approximation.

\emph{Locality-Sensitive Hashing (LSH)} is another paradigm to approximate the maximal inner product $\max_i \omega_i^\top x$, which is known Ma. Following \cite{neyshabur2014symmetric}, we introduce Definition~\ref{def:LSH}.

\begin{definition}\label{def:LSH}
A hash is said to be a $(S, cS, p_1, p_2)$-LSH for a similarity function $\omega^\top x$ over the pair of spaces ${\cal X}, {\cal W} \subseteq {\cal Z}$ if for any $x\in {\cal X}$ and $\omega \in {\cal W}$: 
\begin{itemize}
\item if $\omega^\top x \ge S$ then $\mathbb{P}[h(x) = h(\omega)] \ge p_1$;
\item if $\omega^\top x \le cS$ then $\mathbb{P}[h(x) = h(\omega)] \le p_2$, $0 < c < 1$.
\end{itemize}
\end{definition}

As the optimal value of $\max_k \omega_k^\top x$ is unknown in advance (and might be even  negative), to guarantee the utility of LSH to approximate the margin we require LSH to be universal, i.e. for every $S > 0$ and $0 < c < 1$ it is an $(S, cS)$-LSH \cite{neyshabur2014symmetric}. They also propose the simple LSH algorithm based on random Gaussian projections, with hashing quality 
\begin{gather}
\varrho(c, S) = \log\frac{p_1}{p_2} = \frac{\log \left(1 - \pi^{-1}\cos(S)\right)}{\log \left(1 - \pi^{-1}\cos(cS)\right)}. 
\end{gather}


Following \cite{neyshabur2014symmetric}, one needs $O(n^{\varrho(c, S)} \log 1/\delta)$ to distinguish between $\omega^\top x > S$, and $\omega^\top x < cS$ with probability at least $1-\delta$. Assume below for simplicity that for any $x, \omega: \, \|\omega\|_2 \le 1, \|x\|_2 \le 1$. Applying LSH recursively until 
\[cS > {\varepsilon},\] 
one gets $O(n^{\bar\varrho}\log 1/\delta \log 1/(1-\varepsilon))$ time to construct $(\varepsilon, \delta)$ margin approximation, where 
\[
\bar\varrho = \min_{S \ge \varepsilon, c \ge 1 - \varepsilon} \varrho(c, S) = \min_{S \ge \varepsilon, c \ge 1 - \varepsilon} \log\frac{p_1(c, S)}{p_2(c, S)}.
\]

We also note that the maximal inner product $\max_i \omega_i^\top x$ can be equally stated as a stochastic convex optimization problem:
\begin{gather*}
\mathbb{E}_{j \sim U_d} \sum_{i=1}^c z_i \omega^j_i x^j \to \min_{\substack{\sum_{i=1}^d z_i = 1, z_i \ge 0}},
\end{gather*}
where we use the superscript index to denote a coordinate of the vectors, and denote by $U_d$ the uniform probability measure over $\{1, 2, \dots, d\}$. As it is known from the seminal result of \cite{nemirovski2009robust} on stochastic mirror descent algorithm and \cite{hazan2016variance} on the stochastic Frank-Wolfe there exists potentially sub-linear time complexity algorithms to solve the problem approximately over sparse data. Nevertheless the discussion on optimization approach is out of the scope of this paper. 


\subsection{Multi-Class Support Vector Machines}

Multi-class support vector machines (M-SVMs) still remain top classification methods owing to their accuracy and robustness \cite{rifkin2004defense}. In this section, we analyse simple sub-gradient descent methods to train M-SVMs with $\ell_1-$ and $\ell_2-$ regularization in terms of the influence of inexact margins on their accuracy. Our consideration is mainly inspired by the seminal work of \cite{Wang2010} for support vector machines optimization with inexact oracle. 

\paragraph{$\ell_2$-regularization.} According to \cite{Crammer2001}, the multi-class SVM classifier decision rule is:
\begin{equation*}
    \argmax_{i \in \YY}\omega^T_ix = \argmax_{i \in \YY} \left[\mathbf{W}x\right]_i,
\end{equation*}
where $\omega_i$ is a weight vector of $i$-th classifier and 
\begin{equation*}
    \mathbf{W} := \left[\omega_1, \dots, \omega_C \right] \in \R^{C \times d}
\end{equation*}
is a matrix of weight vectors of all classifiers.

The learning objective is
\begin{equation}
    \label{eq:multiclss_svm_loss}
    L(\mathbf{W}) \doteq \frac{\lambda}{2}\|\mathbf{W}\|^2_F + \frac{1}{n}\sum_{i = 1}^n \ell\left(\mathbf{W}, (x_i, y_i)\right) \to \min_{\mathbf{W}},
\end{equation} 
where $\ell(\mathbf{W}, (x, y))$ is $\rho$-Hinge loss function.

Algorithm \ref{algo:mpegasos} is essentially an extension of the Pegasos algorithm to train the $\ell_2$ regularized support vector machines \cite{Shalev-Shwartz2011}. We further refer to it as MEMOIR-$\ell_2$ The convergence rate of the algorithm is established in Theorem~\ref{thm:pegasos}. 

\begin{theorem}\label{thm:pegasos}
Assume that for all $(x, y) \in S$, $\|x\|_2 \leq 1$. Let $\mathbf{W}^*$ be an optimal solution for the Problem~\eqref{eq:multiclss_svm_loss} and also the batch size $b_t=1$ for all $t$ in Algorithm~\ref{algo:mpegasos}. 
Then for $\bar{\mathbf{W}} = \frac{1}{T} \sum_{t=1}^T \mathbf{W}_t$ one has with probability at least $1-\delta$, $0 < \delta < 1$
\begin{gather*}
f(\bar{\mathbf{W}}) \le f(\mathbf{W}^*) + \frac{c (1 + \log (n)}{2\lambda n} + \psi(\varepsilon, \hat\delta, \lambda), 
\end{gather*}
where on each $i$-th training step we use $(\varepsilon, \hat\delta)$ inexact margin, $\psi\doteq \sqrt{\lambda}^{-1}\min(1, (1-\hat \delta)\varepsilon\sqrt{\lambda} + \hat\delta + \sqrt{\log (1/\delta)/n})$, and $c = \sqrt{\lambda} + 2 + \varepsilon$. 
\end{theorem}
\begin{proof}
Following to \cite{Wang2010} we treat a margin inexactness as an adversarial noise to the gradient and the derive its total influence on the minimization. The bound on $\psi$ is nothing more than application of the Hoeffding inequality to the total distortion introduced by the inexact margin. 
The full proof is provided in the supplementary. 
\end{proof}

Theorem \ref{thm:pegasos} requires the inexactness in margin computation to be bounded by the $\varepsilon_i$ which in its turn should sum up to $o(T)$ for the consistency on the algorithm. This requirement is important from theoretical perspective as it limits the performance of the numerical schemes based on convex optimization and LSH in margin approximation. On the other hand, it is much less crucial from the practical perspective as we see in our numerical study. 




\begin{algorithm}[t]
    \begin{algorithmic}
        \STATE{$\mathbf{W}^0 := 0$}
        \FOR {$t = 1 \dots T$}
            \STATE{Get batch $b_t = \{(x_{j}, y_{j})\}_{j=1}^{|b_t|}$ uniformly at random}
            \STATE{$\mathbf{W}^t := (1-\lambda\eta_t)\mathbf{W}^{t-1}$} 
            \FOR{$(x_{j}, y_{j}) \in b_t$}
                \STATE{Compute approximately \[r_j := \argmax_{r \in \YY \setminus \{y_j\}}{x_j^\top \omega_{r}^{(t)}}\]}
                \IF{$1 + x_j^\top\left(\omega^{(t)}_{r_j} - \omega^{(t)}_{y_j}\right) > 0$}
                    \STATE{$\omega^{(t)}_{r_j} \doteq \omega^{(t)}_{r_j} - \eta_t x_j$}
                    \STATE{$\omega^{(t)}_{y_j} \doteq \omega^{(t)}_{y_j} + \eta_t x_j$}
                \ENDIF
            \ENDFOR
            \STATE{$\phi_t \doteq \min\left\{1, \frac{1}{\sqrt{\lambda}\|W^t\|_2} \right\}$}
            \STATE{$W^{t+1} \doteq \phi_t W^t$}
        \ENDFOR
    \end{algorithmic}
     \caption{$\ell_2$-regularized Multi-class Support Vector Machines with Approximate Maximal Inner Product Search \label{algo:mpegasos}}
\end{algorithm}

\paragraph{$\ell_1$-regularization.}

\begin{algorithm}[t]
    \begin{algorithmic}
        \STATE{$\mathbf{W}^0 := 0$}
        \FOR {$t = 1 \dots T$}
            \STATE{Get batch $b_t = \{(x_j, y_j)\}_{j=1}^{|b_t|}$}
            \STATE{$R_t := \varnothing$}
            \FOR{$(x_j, y_j) \in b_t$}
                \STATE{Compute approximately \[r_j := \argmax_{r \in \YY \setminus \{y_j\}}{x_j^\top w_{r}^{t}}\]}
                \IF{$1 + x_j^\top\left(w^{(t)}_{r_j} - w^{(t)}_{y_j}\right) > 0$}
                    \STATE{$w^{(t)}_{r_j} \doteq w^{(t)}_{r_j} - \eta_t x_j$}
                    \STATE{$w^{(t)}_{y_j} \doteq w^{(t)}_{y_j} + \eta_t x_j$}
                \ENDIF
            \STATE{$R_t = R_t \cup \{y_t, r_t\}$}
            \ENDFOR     
            \FOR{$i \in R_t$}
             \STATE{$w^{(t+1)}_{i} = \text{Truncate}(w^{(t)}_{i}, |R_t|)$  (Eq. \ref{eq:truncate_l1})}
            \ENDFOR
        \ENDFOR
    \end{algorithmic}
     \caption{$\ell_1$-regularized Multi-class Support Vector Machines with Approximate Maximal Inner Product Search \label{algo:lasso_svm}}
\end{algorithm}

In the area of text classification, objects are often described by the TF-IDF or word/collocation frequencies and have sparse representation which is crucial for large-scale machine learning problem. To control the sparsity of $\mathbf{W}$, we use a simple truncated stochastic gradient descent given in Algorithm~\ref{algo:lasso_svm}. It's worth to mention a variety of efficient optimization schemes for $\ell_1$ minimization \cite{jaggi2013revisiting,En-HsuYen2016,goldfarb2017linear,zhu20041}. We believe that similar technique could be utilized for any of the schemes above as well as for the sub-gradient descent we consider here. We also refer to this algorithm as MEMOIR-$\ell_1$.

The problem of $\ell_1$ multi-class support vector machines \cite{zhu20041} is to minimize
\begin{equation}
    \label{eq:l1_svm_loss}
    L(\mathbf{W}) \doteq \frac{\lambda}{2}\|\mathbf{W}\|_1 + \frac{1}{n}\sum_{i = 1}^n \ell\left(\mathbf{W}, (x_i, y_i)\right) \to \min_\mathbf{W}. 
\end{equation} 
A step of stochastic sub-gradient descent method
\begin{equation}
    \omega^{(t+1)}_{i} \doteq \omega^t_i - \eta_t g_i^t,
\end{equation}
where 
\begin{equation*}
        g_i^{t} =   \frac{\lambda}{2}\text{sign}\{\omega^t_i\} + 
        \hspace{-5mm}\Ind{\ell(\omega^t_i, (x_t, y_t)) > 0} \times \begin{cases}
            x_t, & i = r_t \\ 
            -x_t, & i = y_t \\
            0, & \text{otherwise}
        \end{cases}
\end{equation*}
and $\text{sign}\{x\} = \{\text{sign}(x_1), \dots ,  \text{sign}(x_n)\} \in \R^n, \, x \in \R^d$

The details of the method are provided in Algorithm \ref{algo:lasso_svm}. Note an important truncation step which zeroes out sufficiently small elements of $\mathbf{W}$ and significantly reduces memory consumption. We apply it in our algorithm only in the case  of the resulting $\ell_1/\ell_2$ norm of the truncated elements is sufficiently small. In particular, for $\omega \in \R^d$:

\begin{equation}
	\label{eq:truncate_l1}
    \text{Truncate}(\omega, \xi) \doteq [\widetilde{\omega_1}, \dots \widetilde{\omega_d}] \in \R^d,
\end{equation}
where
\[\widetilde{\omega_j} \doteq
\begin{cases}
         \omega_j - \frac{C}{\xi}\lambda\eta_k,  & \text{if } \omega_j> \frac{C}{\xi}\lambda \eta_k \\
         \omega_j + \frac{C}{\xi}\lambda\eta_k, & \text{if }\omega_j < -\frac{C}{\xi}\lambda \eta_k \\
         0, & \text{otherwise.}
     \end{cases}
\]

\begin{table*}[h]
\centering
    \begin{tabular}{|c|c|c|c|c|c|}
        \hline 
        Datasets & \# of Classes, $C$ & Dimension, $d$ & Train Size & Test Size & Heldout Size \\ 
        \hline \hline 
        LSHTC1 & 12294 & 409774 & 126871 & 31718 & 5000 \\
        DMOZ & 27875 & 594158 & 381149 & 95288 & 34506 \\
        WIKI-Small & 36504 & 380078 & 796617 & 199155 & 5000 \\
        WIKI-50k & 50000 & 951558 & 1102754 & 276939 & 5000 \\
        WIKI-100k & 100000 & 1271710 & 2195530 & 550133 & 5000 \\
        \hline
    \end{tabular}
    \caption{\small{Description of datasets used for the numerical evaluation \label{table:datasets}}}
\end{table*}

\begin{theorem}
\label{thm:l1}
After $n$ iterations of Algorithm \ref{algo:lasso_svm} with the full update step and step sizes $\eta_1, \dots, \eta_n$ and corresponding $(\varepsilon, \delta/n)$ inexact margins the expected loss at $\mathbf{\bar W} = \frac{1}{n} \sum_{i=1}^n \mathbf{W}_i$ is bounded with probability $1-\delta$ as
\begin{gather}
f(\mathbf{\bar W}) \le f(\mathbf{W}^*) + \frac{R^2}{2n\eta_n} + \frac{G^2}{2} \sum_{k=1}^n \eta_k + RG \psi 
\end{gather}
where $\psi = 2\varepsilon\sum_{k=1}^n \eta_k$, and $f(\mathbf{W}^*)$ is the optimal value in Eq.~\eqref{eq:l1_svm_loss}, the subgradient's $\ell_2$-norm is bounded by $G$, $\|x_i\|_2 \le R$ for any $i$, and each truncation operation does not change the $\ell_2$ norm of $\mathbf{W}$ on more than $\epsilon$ after each iteration. 
\end{theorem}
\begin{proof}
The proof is a direct implication of the standard sub-gradient convergence analysis to the case of sub-gradient errors due to inexact margin and truncation. The condition on the inexact margin also guarantees that with with probability at least $1-\delta$ each distortion in margin is bounded by $\varepsilon.$
\end{proof}



\section{Numerical Experiments}
In this section we provide an empirical evaluation of the proposed algorithms and compare them to several state-of-the-art approaches. We also discuss how hyperparameter tuning affects algorithms' performance from time, memory and quality prospectives.

\paragraph{Datasets.}
We use datasets from the Large Scale Hierarchical Text Classification Challenge (LSHTC) 1 and 2 \cite{Partalas2015}, which were converted from multi-label to multi-class format by replicating the instances belonging to different class labels. These datasets are provided in a pre-processed format using both stemming and stop-words removal.  Their characteristics, such as train, test, and heldout sizes, are listed in Table \ref{table:datasets}. We would like to thank authors of \cite{Joshi} for providing us with these datasets.

\paragraph{Evaluation Measures.}
During the experiments two quality measures were evaluated: the accuracy and the Macro-Averaged F1 Measure (MaF1). The former represents the fraction of the test data being classified correctly, the later is a harmonic average of macro-precision and macro-recall; the higher values correspond to better performance. Being insensitive to class imbalance, the MaF1 is commonly used for comparing multi-class classification algorithms.

\paragraph{Baselines.}
We compare MEMOIR-$\ell_1$ and $\ell_2$ SVM algorithms with the following multi-class classification algorithms:

\begin{itemize}
    \item \textsc{OVR}: One-vs-rest SVM implementation from LIBLINEAR \cite{Fan2008}.
    \item \textsc{M-SVM}: Multi-class SVM implementation from LIBLINEAR proposed in \cite{Crammer2002}.
    \item \textsc{RecallTree}: A logarithmic time one-vs-some tree-based classifier. It utilises trees for selecting a small subset of labels with high recall and scores them with high precision \cite{Daume2016}.
    \item \textsc{FastXML}: A computationally efficient algorithm for extreme multi-labeling problems. It uses hierarchical partitioning of feature space together with direct optimization of nDCG ranking measure \cite{Prabhu2014}.
    \item \textsc{PfastReXML}: A tree ensemble based algorithm which is an enhanced version of FastXML: the nDCG loss is replaced with its propensity scored variant which is unbiased and assigns higher rewards for accurate tail label predictions \cite{Jain2016}.
    \item \textsc{PD-Sparse}: A recent classifier with a multi-class margin loss and Elastic Net regularisation. To establish an optimal solution the Dual-Space Fully-Corrective Block-Coordinate Frank-Wolfe algorithm is used \cite{En-HsuYen2016}.
\end{itemize}

\paragraph{Hardware Platform.}
In our experiments we use a system with two 20-core Intel(R) Xeon(R) Silver 4114 2.20GHz CPUs and 128 GB of RAM. A maximum of 16 cores are used in each experiment for training and predicting. The proposed algorithms run in a single main thread, but querying and updating ANN search structure is parallelized. We were able to achieve a perfect 8x training time speed up when using 8 cores, 11x speed up when using 16 cores, and 20x speed up when using 32 cores (compared to a single core run of MEMOIR-$\ell_2$ algorithm on LSHTC1 dataset). This means that the proposed algorithm is well-scalable and has reasonable  yet sublinear training time improvement with the number of used cores.

\paragraph{Implementation Details.}
In all algorithms we employ "lazy" matrix scaling, accumulating shared matrix multiplier $acc := acc \cdot \alpha_t$ during all matrix-scalar multiplications in a form $\mathbf{W}^t := \alpha_t \mathbf{W}^t$ and later dividing by $acc$ in addition and subtraction operations when necessary. This allows to keep update time sublinear to problem size. Additionally, we store weight matrix $W$ as $C$ lists of CSR sparse vectors individual to each class, thus requiring $O(nnz(W) + C)$ memory. Finally, in each algorithm we do class prediction for test object $x$ using exact MIPS computation. This is feasible due to the fact that all algorithms achieve highly sparse weight matrices $W$, thus making weight matrix-test object vector multiplication very fast.

%
%

\paragraph{Inexact Margin Computation.}

\begin{table*}[ht]
\centering
    \begin{tabular}{|c|c|c|c|c|c|}
        \hline
        LSHTC1 & $|b| = 0.1\sqrt{C}$ & $|b| = \sqrt{C}$ & $|b| = 10\sqrt{C}$ & $|b| = 100\sqrt{C}$ & $|b| = 1000\sqrt{C}$ \\ 
         & $T = 25000$ & $T = 2500$ & $T = 250$ & $T = 25$ & $T = 3$ \\
        \hline
 Train time  & 1569s & {885s} & 1331s & 1984s & 1830s  \\
 Accuracy & 8.4\% & 26.6\% & 30.8\% & {34.5\%} & 29.3\% \\
 MaF1 & 6.9\% & 19.5\% & 22.7\% & {24.7\%} & 21.0\% \\
        \hline
    \end{tabular}
    \caption{\small{Batch size $|b|$ tuning experiment \label{table:batch_size}}}
\end{table*}

In the case of linear classifiers, the problem of inexact margin computation is known as the Maximum Inner Product Search (MIPS) problem. Many techniques, such as cone trees or locality-sensitive hashing \cite{Neyshabur2015} have been proposed to tackle this problem in a high dimensional setup. We refer to a recent survey \cite{Chen2018} for more details.

To deal effectively in large scale setup, a MIPS solver should maintain the sublinear costs of query and incremental update operations. In practice the majority of solutions lacks theoretical guarantees providing only empirical evaluations \cite{Chen2018}.

In our experiments we use two different MIPS solvers: SimpleLSH, an LSH-based MIPS solver \cite{Neyshabur2015}, and a Navigable Small-World Graph (SW-Graph), a graph-based algorithm introduced in \cite{Malkov2014}. Here we provide theoretical guarantees regarding to the properties of SimpleLSH as an \emph{  $(\varepsilon, \delta)$ inexact} oracle, but our framework can be extended to any similar LSH implementation.

In contrast to SimpleLSH, SW-Graph is a purely empirical algorithm, according to \cite{Malkov2014}, yet we included it into our experiments due to itst high performance in comparison to other solutions \cite{Aumuller2017}. In our implementation we use Non-Metric Space Library (nmslib) \footnote{\url{https://github.com/nmslib/nmslib}} which provides a parallel implementation of SW-Graph algorithm with incremental insertions and deletions. One useful property of nmslib is its ability to work in non-metric similarity functions, including negative dot product, which makes it a suitable implementation of a MIPS oracle.

\subsection{Hyperparameters}

\begin{figure}[t]
\includegraphics[width=\linewidth]{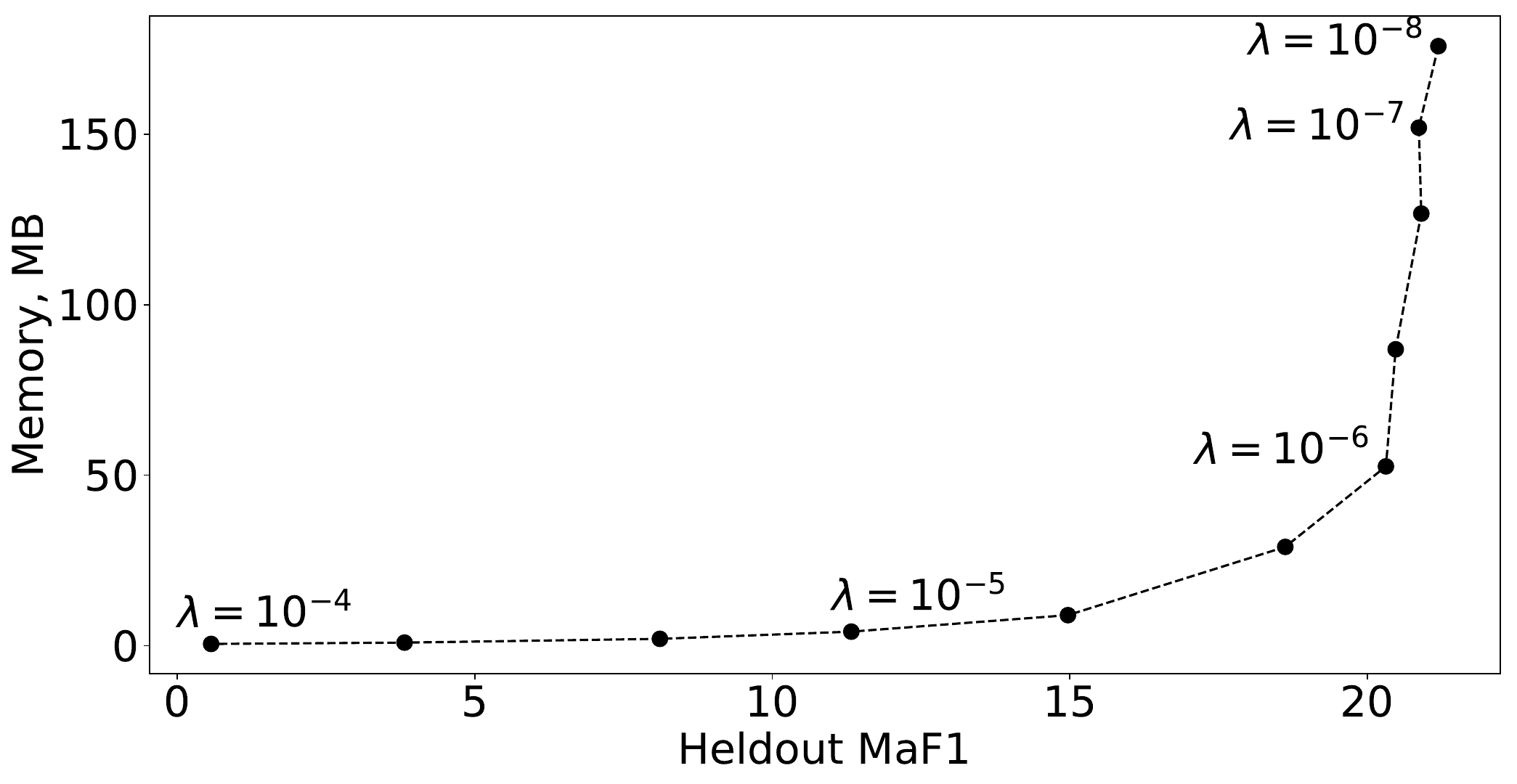}
  \caption{\small{$\ell_1$ regularizer tuning experiment}}
  \label{fig:l1_gamma_tuning}
\end{figure}

In this subsection we discuss reasonable choice of algorithms' parameters and ways how this choice affects training time, memory consumption and quality. The complete list with exact values of all hyperparameters is provided in the Appendix.

\paragraph{Common parameters.}

Learning rate $\eta_t$ is used with $1/t$ learning rate in a form
$\eta_t = \eta_0/(1 + \eta_{step} t)$,
where t is the iteration number and $\eta_0$, $\eta_{step}$ are hyperparameters.

Total number of iterations $T$ were chosen for each dataset independently by obsesving quality improvement on heldout set. When heldout MaF1 does not improve significantly for 5--10 iterations, we stop the learning process.

Batch size $|b|$ was optimized using grid search on logarithmic scale and turned out to be identical for all algorithms and datasets. Generally, increasing batch size $|b|$ while keeping the number of observed objects $|b|T$ fixed first leads to improvement in accuracy at a cost of increased learning time, see Table \ref{table:batch_size} for details.

\paragraph{L2.}
The choice of regularizer's $\lambda$ in case of MEMOIR-$\ell_2$ algorithm affects quality, but has almost no effect on time and memory. This is because $\lambda$ value only affects $W_k$ matrix multiplier and does not have sparsifying effect on it, unlike $\lambda$ in $\ell_1$ SVM. We found $\lambda = 1$ to be a reasonable choice in all our experiments.

\paragraph{L1.}

\begin{figure*}[ht]
\includegraphics[width=\linewidth]{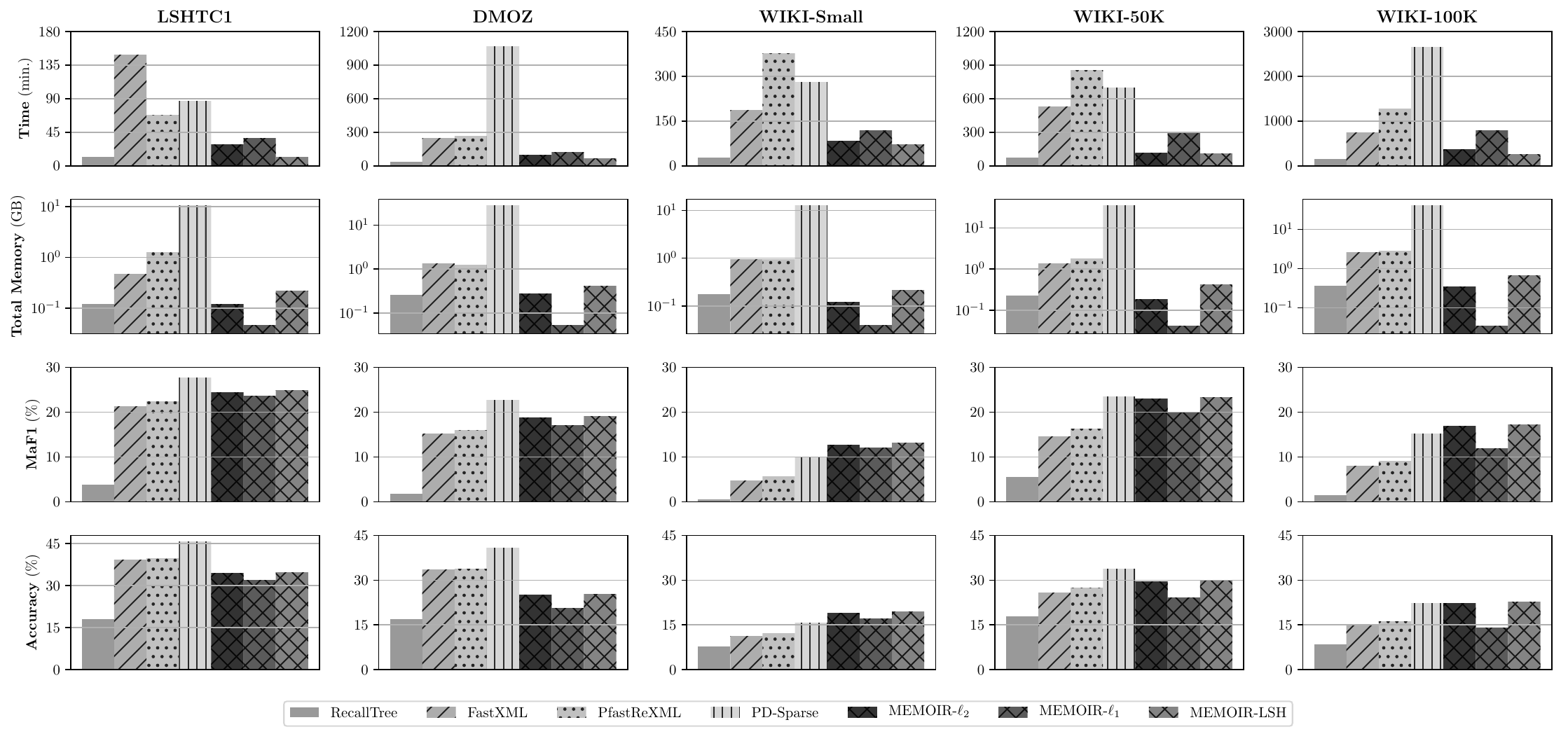}
  \caption{\small{Comparison in time, memory usage, MaF1 and accuracy of the seven best performing methods}}
  \label{fig:pareto_quality}
\end{figure*}

Regularizer's parameter $\lambda$ in MEMOIR-$\ell_1$ algorithm has little effect on training time (it takes only 33\% longer to train MEMOIR-$\ell_1$ with $\lambda=10^{-3}$ compared to $\lambda = 10^{-7}$), but provides a  way to trade quality for memory. Figure \ref{fig:l1_gamma_tuning} illustrates this effect. Decreasing $\lambda$ leads to increasing the number of non-zero elements in the weight matrix, which leads to more accurate predictions made using this weights.

\section{Results}

The results for the baselines and the proposed methods in terms of training and predicting time, total memory usage and predictive performance evaluated with accuracy and MaF1 are provided in Table \ref{table:res_comp}. For a visual comparison we also display the results of seven best models graphically, see Figure \ref{fig:pareto_quality}.

All proposed algorithms consume substantially less memory than the existing solutions with MEMOIR-$\ell_1$ algorithm achieving the most impressive memory reduction. MEMOIR-LSH algorithm achieves the fastest training time (due to reduced feature space and fast computation of Hamming distance) with highest memory usage. MEMOIR-$\ell_2$ is a good compromise between the previous two algorithms in terms of test quality, memory and training time.

\begin{table*}[!h]
\vspace{2mm}
\centering
\scriptsize
 \begin{tabular}{|c|c|c|c|c|c|c|c|c|c|c|}
 \hline
\multirow{2}{*}{Data} & & \multicolumn{6}{|c|}{Baselines} & \multicolumn{3}{|c|}{MEMOIR} \\ \cline{3-11}
    & & \textsc{OVR} & \textsc{M-SVM}  & \textsc{RecallTree} & \textsc{FastXML} & \textsc{PfastReXML} & \textsc{PD-Sparse} & \textsc{$\ell_2$} & \textsc{$\ell_1$} & \textsc{LSH} \\
 \hline\hline

 {LSHTC1} & Train time  & 23056s & 48313s & 701s & 8564s & 3912s & 5105s & 1673s & 2204s & {655s} \\
 n = 163589      & Predict time& 328s & 314s &  {21s} & 339s   & 164s & 67s & 18s & {6s} & 12s \\
 d = 409774      & Total memory& 40.3G & 40.3G &  {122M} & 470M   & 471M & 10.5G & 119M & {46M} & 218M \\
 C = 12294       & Accuracy       & 44.1\% & 36.4\%& 18.1\%& 39.3\% & 39.8\% & {45.7\%} & 34.5\% & 31.9\% & 34.6\% \\
    & MaF1        & 27.4\% & 18.8\% & 3.8\% & 21.3\% & 22.4\% & {27.7\%} & 24.4\% & 23.6\% & 24.8\% \\ \hline

{DMOZ}   & Train time   & 180361s & 212356s & {2212s} & 14334s & 15492s & 63286s & 5709s & 7226s & {3637s} \\
n = 510943      & Predict time & 2797s   & 3981s &  {47s} & 424s & 505s & 482s & 76s & {22s} & 77s \\
d = 594158      & Total memory & 131.9G  & 131.9G &  {256M} & 1339M & 1242M & 28.1G & 271M & {52M} & 417M \\
C = 27875 & Accuracy       & 37.7\% &32.2\%  & 16.9\% & 33.4\% & 33.7\% & {40.8\%} & 25.1\% & 20.6\% & 25.2\% \\
& MaF1          & 22.2\%  & 14.3\% & 1.8\% & 15.1\% & 15.9\% & {22.7\%} & 18.8\% & 17.1\% & 19.1\% \\\hline

 {WIKI-Small} & Train time    & 212438s & $>$4d & {1610s} & 10646s & 21702s & 16309s & 4791s & 7055s & 4165s \\
n = 1000772          & Predict time & 2270s & NA & {24s} & 453s  & 871s & 382s & 88s & 36s & 61s \\
d = 380078           & Total memory & 109.1G & 109.1G &  {178M} & 949M & 947M & 12.4G & 121M & {39M} & {213M} \\
C = 36504            & Accuracy       & 15.6\% & NA & 7.9\% & 11.1\% & 12.1\% & 15.6\% & {19.0\%} & 17.0\% & {19.4}\% \\
& MaF1          & 8.8\% & NA & $<$1\% & 4.6\% & 5.6\% & 9.9\%  & {12.7\%} & 12.1\% & {13.1}\% \\\hline

{WIKI-50K} & Train time     & NA   & NA  & {4188s} & 30459s & 48739s & 41091s & 6755s & 17303s & {6215s}  \\
n = 1384693         & Predict time & NA   & NA  & {45s} & 1110s  & 2461s  & 790s & 120s & 59s & 120s  \\
d = 951558          & Total memory & 330G & 330G&  {226M} & 1327M  & 1781M  & 35G & 185M & {42M} & 424M    \\
C = 50000           & Accuracy       & NA & NA & 17.9\% & 25.8\% & 27.3\% & {33.8\%} & 29.6\% & 24.2\% & 30.0\%  \\
& MaF1          & NA   & NA  & 5.5\% & 14.6\% & 16.3\% & {23.4\%} & 22.9\% & 20.0\% & 23.3\%  \\ \hline

{WIKI-100K} & Train time     & NA   & NA   & {8593s} & 42359s & 73371s & 155633s & 21061s & 46730s & {14323s}  \\
n = 2750663          & Predict time & NA   & NA   & {90s} & 1687s  & 3210s  & 3121s & 457s & 161s & 504s \\
d = 1271710          & Total memory & 1017G& 1017G&  {370M} & 2622M  & 2834M  & 40.3G & 346M & {35M} & 660M   \\
C = 100000           & Accuracy       & NA & NA & 8.4\% & 15.0\% & 16.1\% & 22.2\% & {22.3\%} & 14.0\% & {22.6\%} \\
& MaF1         & NA   & NA   & 1.4\% & 8.0\% & 9.0\% & 15.1\% & {16.9\%} & 11.8\% & {17.2\%}  \\ \hline

 \end{tabular}

 \caption{\small{Comparison of the result of various baselines in terms of time, memory, accuracy, and macro~F1-measure}}
\label{table:res_comp}
\end{table*}

\section{Conclusion}
Our paper is about extreme multi-class classification using inexact margin computation. 
We provided theoretical analysis of a classification risk in multi-class (one-vs-rest and Crammer-Singer) settings and showed that inexact margin computation does not lead to a significant loss of models' risk. 
We then designed three efficient methods, MEMOIR-$\ell_2$, MEMOIR-$\ell_1$ and MEMOIR-LSH, that solve Crammer-Singer multi-class SVM problem with inexact approximation of a margin using two different MIPS solvers, SW-Graph and SimpleLSH. 
We illustrated an empirical performance of these methods on five extreme classification datatets, on which we achieved good results in terms of quality, memory and training time. 

Finally, we discussed how parameter tuning affects algorithms' performance and provided a practical advice on how to choose hyperparameters for the proposed algorithms. Our implementation is publicly available in a form of an open-source library.


\bibliography{references}
\bibliographystyle{unsrt}


\newpage\newpage
\section*{Appendix for \\ MEMOIR: Multi-class Extreme Classification with Inexact Margin}
\subsection{Proof of Theorem 2}

\setcounter{theorem}{1}
\setcounter{equation}{0}

\begin{theorem}\label{thm:pegasos}
Assume that for all $(x, y) \in S$, $\|x\|_2 \leq 1$. Let $\mathbf{W}^*$ be an optimal solution for the Problem~\eqref{eq:multiclss_svm_loss} and also the batch size $b_t=1$ for all $t$ in Algorithm~\ref{algo:mpegasos}. 
Then for $\bar{\mathbf{W}} = \frac{1}{T} \sum_{t=1}^T \mathbf{W}_t$ one has with probability at least $1-\delta$, $0 < \delta < 1$
\begin{gather*}
f(\bar{\mathbf{W}}) \le f(\mathbf{W}^*) + \frac{c (1 + \log (n)}{2\lambda n} + \psi(\varepsilon, \hat\delta, \lambda), 
\end{gather*}
where on each $i$-th training step we use $(\varepsilon, \hat\delta)$ inexact margin, $\psi\doteq \sqrt{\lambda}^{-1}\min(1, (1-\hat \delta)\varepsilon\sqrt{\lambda} + \hat\delta + \sqrt{\log (1/\delta)/n})$, and $c = \sqrt{\lambda} + 2 + \varepsilon$. 
\end{theorem}
\begin{proof}
Following to \cite{Wang2010} will treat the difference between the exact margin and the inexact one as the gradient error. 

For the convenience we refer as 
\[f_t(\mathbf{W}) = \lambda \|\mathbf{W}\|^2_2/2 + \ell(\mathbf{W}, (x^t, y^t))\] 
the instantaneous loss on $t$-th object. 
Thus 
\[
\mathbf{W}^{t+1} = \Pi_C (\mathbf{W}^t - \eta_t \mathbf{\bar \nabla}^t), \; 
{\bar \nabla}^t = \mathbf{\nabla}^t + {\Delta}^t
\]
where $C$ is a normed ball, $\|W\|_2 \le 1$ according to Lemma 1 of \cite{Shalev-Shwartz2011} and Lemma 1 of \cite{Wang2010}. The distortion $\Delta_t$ bounded as $\|\Delta_t\|_2 \le \varepsilon$ with probability at least $1-\delta$ by the definition of inexact margin. 

The relative progress towards the optimal solution $W^*$ at $t$-th iteration is 
\begin{align}\label{eq:peg}
D_t = 	& \|\mathbf{W}^t - \mathbf{W}^*\|_2^2 - \|\mathbf{W}^{t+1} - \mathbf{W}^*\|_2^2 \ge \nonumber\\
		&\|\mathbf{W}^t - \mathbf{W}^*\|_2^2 - \|\Pi_C(\mathbf{W}^t - \eta_t {\bar \nabla}_t) - \mathbf{W}^*\|_2^2 \ge \nonumber\\
		& \|\mathbf{W}^t - \mathbf{W}^*\|_2^2 - \|\mathbf{W}^t - \eta_t {\bar \nabla}_t) - \mathbf{W}^*\|_2^2 = \nonumber\\
		&-\eta^2_t \|{\bar \nabla}_t\|_2^2 + 2\eta_t (\mathbf{W}^t - \mathbf{W}^*)^\top \nabla^t + \nonumber\\ 
        & \hspace{34mm} 2 \eta_t (\mathbf{W}^t - \mathbf{W}^*)^\top \Delta^t \ge \nonumber\\ 
& - \eta_t^2 G^2 + 2 \eta_t (f_t(\mathbf{W}^t) - f_t(\mathbf{W}^*) + \nonumber\\
& \hspace{25mm} \frac{\lambda}{2} \|\mathbf{W}^t - \mathbf{W}^*\|_2^2) - 4 \eta_t \frac{\|\Delta^t\|_2}{\sqrt{\lambda}}
\end{align}
where the last inequality is due to the contraction property of projection on a convex set, and $\|{\bar \nabla}^t\| \le \lambda \|W^t\|_2 + 2 + \|\Delta^t\| \le \sqrt{\lambda} + 2 + \varepsilon$. 

By the strong convexity of $f_t(W)$ we have
\[
(\mathbf{W}^t - \mathbf{W}^*)^\top\nabla^t \ge f_t(\mathbf{W}^t) - f_t(\mathbf{W}^*) + \lambda \frac{\|\mathbf{W}^t - \mathbf{W}^*\|_2^2}{2}.
\]

As $\nabla_t = \nabla_{\mathbf{W}} f_t(\mathbf{W})$ and the strong convexity of the objective we have $\|\mathbf{W}\|_2^2 \le 2/\sqrt{\lambda}$. Dividing both sides of~Eq.~\eqref{eq:peg} by $2\eta_t$ and rearranging we get: 
\[f_t(\mathbf{W}^t) - f_t(\mathbf{W}^*) \le \frac{D_t}{2\eta_t} - \frac{\lambda \|\mathbf{W}^t - \mathbf{W}^*\|_2^2}{2} + \frac{\eta_t G^2}{2} + 2\frac{\|\Delta^t\|_2}{\sqrt{\lambda}}. 
\]

Summing over all $t$ one has 
\begin{align}\label{eq:sum_peg}
\sum_{t=1}^n f_t(\mathbf{W}^t) & - \sum_{t=1}^n f_t(\mathbf{W}^*) \le \sum_{t=1}^n \frac{D_t}{2\eta_t} + \frac{G^2}{2} \sum_{t=1}^n \eta_t \nonumber\\
&   - \sum_{t=1}^n \frac{\lambda}{2} \|\mathbf{W}^t - \mathbf{W}^*\|_2^2  + \frac{2}{\sqrt{\lambda}} \sum_{t=1}^n \|\Delta^t\|_2 
\end{align}

Rearranging the the terms and using that $\eta_t = 1/(\lambda t)$ we have 
\begin{align}\label{eq:cancel}
\frac{1}{2}\sum_{t=1}^n &\left(\frac{D_t}{\eta_t} - \lambda \|\mathbf{W}^t - \mathbf{W}^*\|_2^2\right) = \nonumber\\
& \hspace{20mm} - \frac{1}{2\eta_n} \|\mathbf{W}^{n+1} - \mathbf{W}^*\|_2^2 \le 0
\end{align}

Finally combining the inequalities \eqref{eq:cancel}, \eqref{eq:sum_peg}, \eqref{eq:peg} and Jensen's inequality we get 
\begin{gather*}
f(\bar{\mathbf{W}}) \le f(\mathbf{W}^*) + \frac{c (1 + \log n)}{2\lambda n} + 2\sum_{i=1}^n \frac{\|\Delta^t\|_2}{\sqrt{\lambda} n}, 
\end{gather*}
where each $\Delta^t$ is bounded as $\|\Delta^t\| \le \varepsilon$ with probability at least $1-\hat \delta$, and always bounded as $\|\Delta^t\| \le 1/\sqrt{\lambda}$ due to the bound on $\|W\|_2$. Thus with probability at least $1-\delta$ by the Hoeffding inequality 
\[
	\sum_{i=1}^n \frac{\|\Delta^t\|_2}{\sqrt{\lambda} n} \le \min\left\{\frac{1}{\sqrt{\lambda}}, (1 - {\hat \delta})\varepsilon + \frac{\hat \delta}{\sqrt{\lambda}} + \sqrt{\frac{\log 1/\delta}{n\lambda}}\right\},
\]
which finishes the proof of the theorem. 
\end{proof}

\subsection{Parameters in Numerical Study}
\begin{table*}[!htb]
    \centering
    \resizebox{\textwidth}{!}{
    \begin{tabular}{|c|c|c|c|c|c|c|}
        \hline
       Algorithm & Parameters  & LSHTC1 & DMOZ & WIKI-Small & WIKI-50K & WIKI-100K \\ \hline \hline
       $ \textsc{OVR} $ &  C & 10  & 10 & 1 & NA & NA  \\ \hline
       $ \textsc{M-SVM} $ &  C & 1 & 1& NA & NA & NA \\ \hline
       $\textsc{RecallTree} $ & --b & 30 & 30 & 30& 30 & 28 \\ \cline{2-7}
        & --l   &1 &0.7 & 0.7& 0.5& 0.5\\ \cline{2-7}
        & --loss\_function  & Hinge& Hinge& Logistic &Hinge &Hinge \\ \cline{2-7}
        & --passes & 5 & 5& 5& 5& 5 \\ \hline
        $ \textsc{FastXML} $ & -t & 100 & 50 & 50 & 100 &50 \\ \cline{2-7}
                   & -c & 100 & 100 & 10 & 10 & 10\\ \hline
         $ \textsc{PfastReXML} $ & -t &50  &50 & 100&200 &100 \\ \cline{2-7}
                   & -c & 100 & 100 & 10 &10 &10 \\ \hline
        $ \textsc{PD-Sparse} $ &  -l & 0.01 & 0.01& 0.001& 0.0001& 0.01\\ \cline{2-7}
         & Hashing & multiTrainHash & multiTrainHash& multiTrainHash& multiTrainHash& multiTrainHash\\ \hline \hline
       $ \textsc{MEMOIR-*} $ & $\eta_0$ & 0.1 & 0.1 & 0.1 & 0.1 & 0.1  \\ \cline{2-7}
       & $\eta_{step}$ & 0.02 & 0.02 & 0.02 & 0.02 & 0.02  \\ \cline{2-7}
       &  $|b|$ & $100\sqrt{C}$  & $100\sqrt{C}$ & $100\sqrt{C}$ & $100\sqrt{C}$ & $100\sqrt{C}$  \\ \hline
       $ \textsc{MEMOIR-}\ell_2 $ & $\lambda$ & 1 & 1 & 1 & 1 & 1  \\ \cline{2-7}
       & $T$ & 25 & 40 & 48 & 60 & 80  \\ \hline
       $ \textsc{MEMOIR-}\ell_1 $ & $\lambda$ & $10^{-6}$ & $10^{-6}$ & $10^{-6}$ & $10^{-6}$ & $10^{-6}$  \\ \cline{2-7}
       & $T$ & 25 & 40 & 48 & 60 & 80  \\ \hline
       $ \textsc{MEMOIR-LSH} $ & $\lambda$ & 1 & 1 & 1 & 1 & 1  \\ \cline{2-7}
       & $T$ & 25 & 40 & 48 & 60 & 80  \\ \hline
       & hash string length & 64 & 64 & 64 & 64 & 64  \\ \hline

    \end{tabular}
    }
\caption{\small{Hyper-parameters used in the final experiments}}
    \label{tab:hyperpms}
\end{table*}

\end{document}